\documentclass{article}

 \usepackage[main, final]{neurips_2025}

\usepackage[table,xcdraw]{xcolor}
\usepackage{tcolorbox}
\usepackage{enumitem}
\usepackage{graphicx}
\usepackage[T1]{fontenc}    
\usepackage{hyperref}       
\usepackage{url}            
\usepackage{booktabs}       
\usepackage{amsfonts}       
\usepackage{nicefrac}       
\usepackage{microtype}      
\usepackage{amsmath}
\usepackage{wrapfig}
\usepackage{multirow}

\definecolor{lightrow}{gray}{0.93}   
\definecolor{mygray}{gray}{0.60}     
\definecolor{darkred}{RGB}{220,0,0}
\definecolor{darkgreen}{RGB}{0,100,0}

\newcommand{\inc}[1]{$_{\scriptstyle\textbf{\textcolor{darkred}{(+#1)}}}$}
\newcommand{\dec}[1]{$_{\scriptstyle\textbf{\textcolor{darkgreen}{(-#1)}}}$}

\title{WebSynthesis: World-Model-Guided MCTS for Efficient WebUI-Trajectory Synthesis}

\author{%
  Yifei Gao, \; Junhong Ye, \; Jiaqi Wang \; and \; Jitao Sang\thanks{Corresponding Authors} \\
  Beijing Jiaotong University\\
  Beijing, China \\
  \texttt{\{yifeigao, jtsang\}@bjtu.edu.cn}\\
  \texttt{\textbf{Project Link}:\href{https://github.com/LucusFigoGao/WebSynthesis}{https://github.com/LucusFigoGao/WebSynthesis}}
}

\begin{document}

\maketitle

\begin{abstract}
    Recent advancements in large language models (LLMs) have significantly improved the capabilities of web agents. However, effectively navigating complex and dynamic web environments still requires more advanced trajectory-level planning and execution. Prior studies have addressed self-improving agents by collecting extensive GUI trajectories from real-environment interactions. Despite their effectiveness, these approaches encounter two critical challenges: (1) \textbf{Uncontrollable environment states}, where real or sandboxed web environments often yield unstable and non-deterministic feedback, complicating the reproduction and debugging of agent behaviors; and (2) \textbf{High API costs}, as generating even a single interaction trajectory can involve hundreds of queries, leading to considerable API usage and computational expenses. To address these limitations and enable scalable self-improvement for agents, we propose \textbf{WebSynthesis}, a novel framework for trajectory synthesis and training. WebSynthesis leverages a learned world model to simulate virtual web environments, allowing a policy agent to perform efficient and reversible tree-based planning. This approach supports the large-scale generation of diverse and high-quality trajectories, which are subsequently utilized to refine the agent's policy. Experimental results demonstrate that an agent trained using WebSynthesis on a small-scale synthetic dataset achieves performance comparable to or even surpassing that of models trained on large-scale real-world data.
    

\end{abstract}

\section{Introduction}\label{sec-intro}
Automated agents powered by large language models (LLMs) have demonstrated remarkable potential for web navigation tasks~\cite{survey1, survey2, WebPilot, AgentOccam, OS-Genesis, WebEvolver, AutoWebGLM, WebAgent-R1}. Web agents, such as Operator (OpenAI), Manus (Manus AI), are expected as human-like "eyes and hands," interacting with real-world web environments through multi-turn perception (reading) and action (writing) until successfully completing user-defined tasks. Consequently, training these agents on high-quality trajectories is crucial for enhancing their agentic capabilities \cite{Synapse, OS-Genesis, AgentTrek}. However, gathering large-scale, multi-step trajectories for web navigation tasks remains challenging. The dynamic and diverse nature of web interfaces makes manual demonstration collection both labor-intensive and unsustainable~\cite{WebDreamer, WMA, AgentTrek, UI-TARS}. Additionally, directly training agents via live web interactions is often unpredictable and costly due to noise from real-world environments and the substantial number of API calls required, making algorithm like reinforcement learning (RL) prohibitively expensive~\cite{webRL, AutoWebGLM, WebAgent-R1}.

To address this data bottleneck, recent studies have increasingly investigated trajectory-level self-evolution approaches, where agents iteratively improve by learning from autonomously generated trajectories~\cite{OS-Genesis, PAE, AgentTrek, WebEvolver}. This paradigm allows agents to bootstrap their training data, reducing dependence on human annotations and potentially uncovering novel interaction patterns beyond those demonstrated by humans. While effective, significant challenges persist in existing trajectory synthesis methods. First, accurately modeling the full complexity of modern web environments remains inherently difficult~\cite{WebArena, VisualWebArena}. Even advanced agents may fail when interacting with imperfect simulators, leading to limited scenario coverage or unrealistic behavior. Specifically: (1) Purely self-directed exploration often generates trajectories lacking diversity, causing learning to stagnate as agents repeatedly encounter familiar patterns~\cite{PAE}. (2) Rule-based or tutorial-guided approaches typically address only predefined task templates, leaving many edge cases unexplored~\cite{AgentTrek}. Another critical issue is the high computational cost of data generation~\cite{Agent-Q}, as some synthesis pipelines require thousands of interactions per trajectory, resulting in millions of inference tokens and rendering the approach computationally prohibitive.

Motivated by these limitations, we propose \textbf{WebSynthesis}, a novel framework integrating world modeling~\cite{WMA, WebDreamer, WebEvolver} with search-based trajectory generation mechanisms~\cite{Tree-Search, WebPilot}. Acting as a high-fidelity proxy for the real web, world model facilitates the generation of diverse and rich trajectories without incurring the high costs associated with live interactions, thereby providing an imagined environment for effective agent exploration. To improve the quality and task relevance of synthesized trajectories of policy agents in the world model environment, we employ Monte Carlo Tree Search (MCTS)~\cite{MCTS-survey} to guide the synthesis towards a specific goal. This approach enables the world model to produce interaction sequences tailored to a wide variety of tasks and interface conditions, ensuring trajectories are goal-directed and diverse. Crucially, the search evaluates multiple hypothetical action paths within the world model, selectively retaining those leading to coherent and successful outcomes. By combining model-based environment simulation with goal-oriented search, WebSynthesis efficiently generates controllable and cost-effective web navigation trajectories.

During training, we adopt a two-stage curriculum learning framework. The first stage focuses on strengthening the  fundamental UI understanding capabilities of policy agent, including tasks such as state captioning, functionality description, and state transition prediction ~\cite{UI-TARS, InfiGUIAgent}. This prepares the agent to quickly adapt to complex and unfamiliar web interfaces. In the second stage, we perform Supervised Fine-Tuning (SFT) using the synthesized trajectory data. These synthetic trajectories enable the policy agent to imagine and rehearse a wide range of tasks within a safe virtual environment, thereby improving real-world generalization and accelerating self-improvement beyond the data limitations of prior methods. Experimental results demonstrate that with about 4k synthetic trajectories, WebSynthesis achieves a 20.15\% overall Pass@3 success rate, outperforming OS-Genesis-7B\cite{OS-Genesis} (18.66\%, trained on 7.4k real-world trajectories) and AgentTrek-7B~\cite{AgentTrek} (11.94\%, trained on 20k tutorial-guided synthetic trajectories). Additionally, a TextUI warm-up study reveals that pretraining the three UI fundamental capabilities prior to trajectory-level fine-tuning boosts OS-Genesis performance by +33.4\%, highlighting the critical role of UI understanding in agent training.

Our primary contributions are as follows: (1) We propose WebSynthesis, a novel framework that integrates an LLM-based world model with MCTS to synthesize diverse and controllable web interaction trajectories entirely offline. (2) We design a two-stage curriculum learning paradigm to comprehensively improve the UI fundamental understanding and web navigation capabilities. (3) With only ~4,000 synthetic samples, WebSynthesis outperforms methods trained on significantly larger datasets, including both real-world-collected and tutorial-guided synthetic trajectories.


\section{Related Works}\label{sec-related}

\subsection{Self-Evolving Agents}
Self-evolving agents primarily improve trajectory quality and generalization capabilities through reinforcement learning or synthetic data generation.

\textbf{Reinforcement learning approaches}. AutoWebGLM \cite{AutoWebGLM} leverages carefully curated web browsing data to bootstrap the model through a three-stage curriculum learning framework, where the latter two stages incorporate Direct Preference Optimization (DPO) and rejection sampling for online scenarios. Approaches like WebRL \cite{webRL} introduces a self-evolving online curriculum that automatically generates new tasks from the agent’s failures and employs an outcome-based reward model, enabling an open-source agent to steadily improve and, in some cases, match or surpass GPT-4 on specific benchmarks. WebAgent-R1 \cite{WebAgent-R1}, in contrast, demonstrates that a simple end-to-end reinforcement learning setup, combined with asynchronous trajectory exploration and binary success feedback, can significantly enhance long-horizon task success for open LLM agents, highlighting the effectiveness of direct environment interaction.

\textbf{Data synthesis approaches}. Complementing the above RL approaches, several studies address data scarcity by synthesizing rich training trajectories. OS-Genesis \cite{OS-Genesis} reverses the conventional data collection process: the agent first explores GUI environments freely, and retrospective analysis then extracts high-quality tasks from its trajectories using a reward model. This approach generates diverse and realistic data without the need for human scripting. PAE \cite{PAE} adopts a similar strategy for autonomous skill discovery, enabling the agent to propose new web tasks and self-evaluate outcomes using a vision-language model. This allows the agent to construct its own curriculum and reward signal, enabling policy refinement beyond any fixed instruction set. Meanwhile, AgentTrek \cite{AgentTrek} leverages publicly available web tutorials as surrogates for human demonstrations. It converts instructional content into step-by-step actions and validates them through a vision-language agent, thereby scaling multimodal trajectory generation at low cost and enhancing generalization across both textual and visual web benchmarks. WebCoT \cite{WebCoT} identifies essential reasoning skills for effective web agents, including reflection, branching, and rollback, and synthesizes the corresponding reasoning trajectories.

\subsection{World Model}
In the era of large language models (LLMs), world models have emerged as generative AI systems designed to capture the dynamics of the real world, including its physical and spatial properties ~\cite{world-model-survey1, world-model-survey2}. In RL, world models are used to simulate future observations and environment feedback, enabling policy agents to learn and plan without interacting directly with the real environment \cite{world-model-survey3}. Recent work in web navigation adopts a "simulate-before-act" paradigm \cite{Simulate_Before_Act}. Approaches such as WebDreamer \cite{WebDreamer} and WMA \cite{WMA} leverage LLMs as world models to predict the outcomes of candidate actions in advance, thereby avoiding irreversible errors and reducing the cost of real-time interactions. However, the effectiveness of these methods is heavily dependent on the realism of the world model, which limits their reliability for online planning. WebEvolver \cite{WebEvolver} extends this paradigm by co-evolving the agent and the world model, allowing the agent to improve its decision-making by planning within an increasingly accurate simulated environment. Our work builds on this line of research by integrating LLM-based world modeling with goal-directed search. The world model is trained to generate detailed next-page observations, such as the DOM or accessibility tree, enabling the policy agent to perform efficient and diverse web navigation within a virtual environment.

\section{Methodology}\label{sec3}
In this section, we present the details of the WebSynthesis pipeline, which contains both the data collection process and the curriculum learning framework.

\textbf{Formulations}. Following the standard reinforcement learning framework, we formulate web navigation as a partially observable Markov decision process (POMDP), consisting of four main components: the observation space ($\mathcal{O}$), action space ($\mathcal{A}$), transition function ($\mathcal{T}$), and reward function ($\mathcal{R}$). In our approach, a web world simulator serves as the transition function $\mathcal{T}$, mapping a state-action pair $(\mathcal{O}, \mathcal{A})$ to the next state observation. The policy agent $\pi$ processes a user query $\text{q}$ and engages in multi-step interaction within the world model environment, rather than operating on the real web interface. Consistent with prior work, we represent web page observations using the accessibility tree (A11y), which captures a structured set of accessibility-related states and properties.

\subsection{Stage 1: UI Fundamental Understanding}\label{sec3.1}
\begin{figure*}[!tp]
    \centering
    \includegraphics[width=1\linewidth]{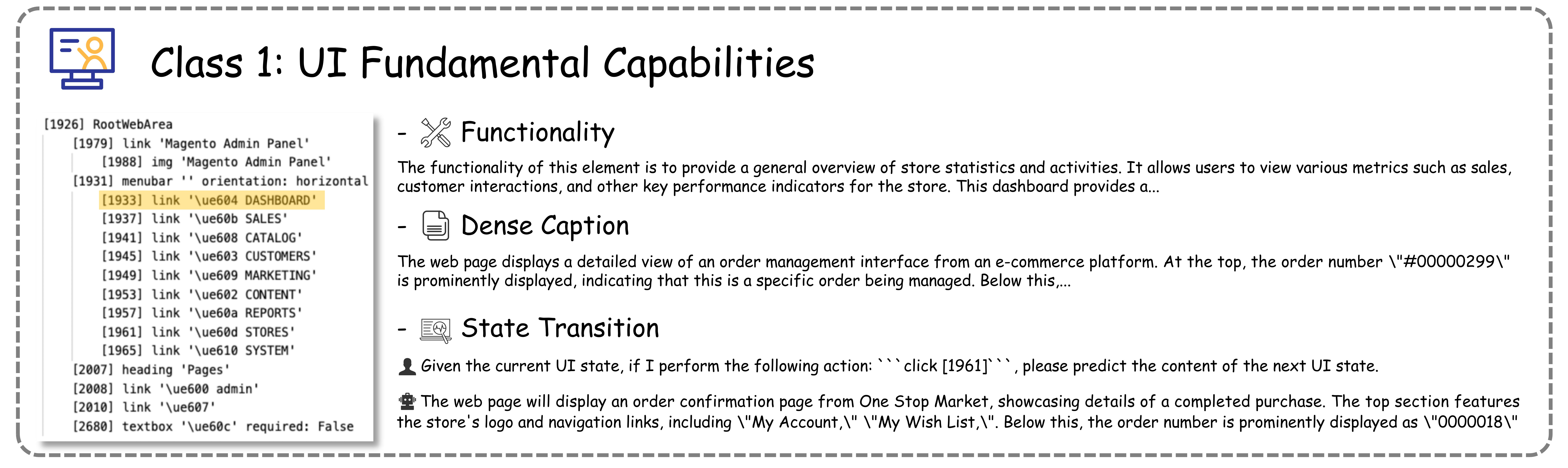}
    \caption{An overview of the UI Fundamental Understanding datasets, which offer single-step, fine-grained annotations, including dense captions, element functionality, and state transitions, designed to train the model to understand, describe, and predict web page states.}
    \label{fig:class1}
\end{figure*}

In headless browsing settings, the policy agent primarily interacts with text-based page representations, which may involve various data formats such as HTML, the DOM tree, or the accessibility tree (A11y). Given the complexity of TextUI scenarios, even when complete page information is provided, the agent may struggle to fully comprehend the contextual information. To address this issue, we first collected and synthesized a series of TextUI datasets for the initial stage of supervised fine-tuning. We primarily adopt the public WebArena environment \cite{WebArena}, a dynamic GUI sandbox that provides both textual observations (e.g., HTML and A11y) and visual observations (e.g., screenshots) of web pages. By performing random exploration in this environment, we obtain a large number of transition triples $(o_{t-1}, a_{t}, o_{t})$, representing the transition from observation $o_{t-1}$ to $o_{t}$ given action $a_{t}$. Based on the collected transition triples, we follow the task definitions in \cite{UI-TARS} and curate three core capabilities for TextUI understanding (see Figure \ref{fig:class1} for illustration): (1) dense captioning, (2) element functionality prediction, and (3) state transition captioning. The overall instruction template are shown in Table \ref{tab:textui_caption} \ref{tab:textui_function} \ref{tab:textui_transmission}.

\textbf{Dense Captioning}: To enhance the model’s global understanding of TextUI inputs, we address a key limitation of text-based representations, the absence of visual layout information. Directly summarizing textual content often proves insufficient for capturing the structure and context of complex interfaces. To overcome this, we leverage GPT-4o by providing it with corresponding GUI screenshots, enabling the generation of comprehensive and detailed descriptions that capture not only individual UI elements but also their spatial relationships and overall layout. During training, these descriptions are paired with text-based observations and provided as input to the model.

\textbf{Element Functionality}: To improve the recognition and understanding of specific elements in TextUIs (e.g., "textbars" or "clickable elements"), we focus on generating detailed, structured descriptions for each interactive component. Leveraging GPT-4o’s strong visual understanding capabilities, similar to SoM~\cite{SoM}, we prompt the model to synthesize functional descriptions based on both the text-based representation and the corresponding GUI screenshot. These synthesized descriptions are then paired with the text observation (i.e., A11y), replacing the screenshot as contextual input to form a QA pair.

To mitigate the generation of redundant QA pairs resulting from the excessive length of A11y, we apply a local compression strategy during preprocessing. Specifically, we retain only the elements adjacent to the target UI element, while preserving the original hierarchical structure of the A11y tree. This approach reduces unnecessary context and significantly enhances the diversity and efficiency of the resulting dataset.

\textbf{State Transition Prediction}: To further enhance the model’s understanding of UI elements, particularly in response to user interaction requests, we aim to equip the model with the ability to predict changes in the main content of the page. To achieve this, we capture the differences between consecutive observations and incorporate them into the agent’s reasoning process. During training, the agent is provided with the current page observation and the corresponding action, and is tasked with generating the changes of layout and a caption for the subsequent page frame. This capability is essential for tasks requiring fine-grained interaction understanding and dynamic state awareness.


\subsection{Stage 2: A World Model Guided Monte Carlo Tree Search}\label{sec3.2}
\begin{figure*}[t]
    \centering
    \includegraphics[width=1\linewidth]{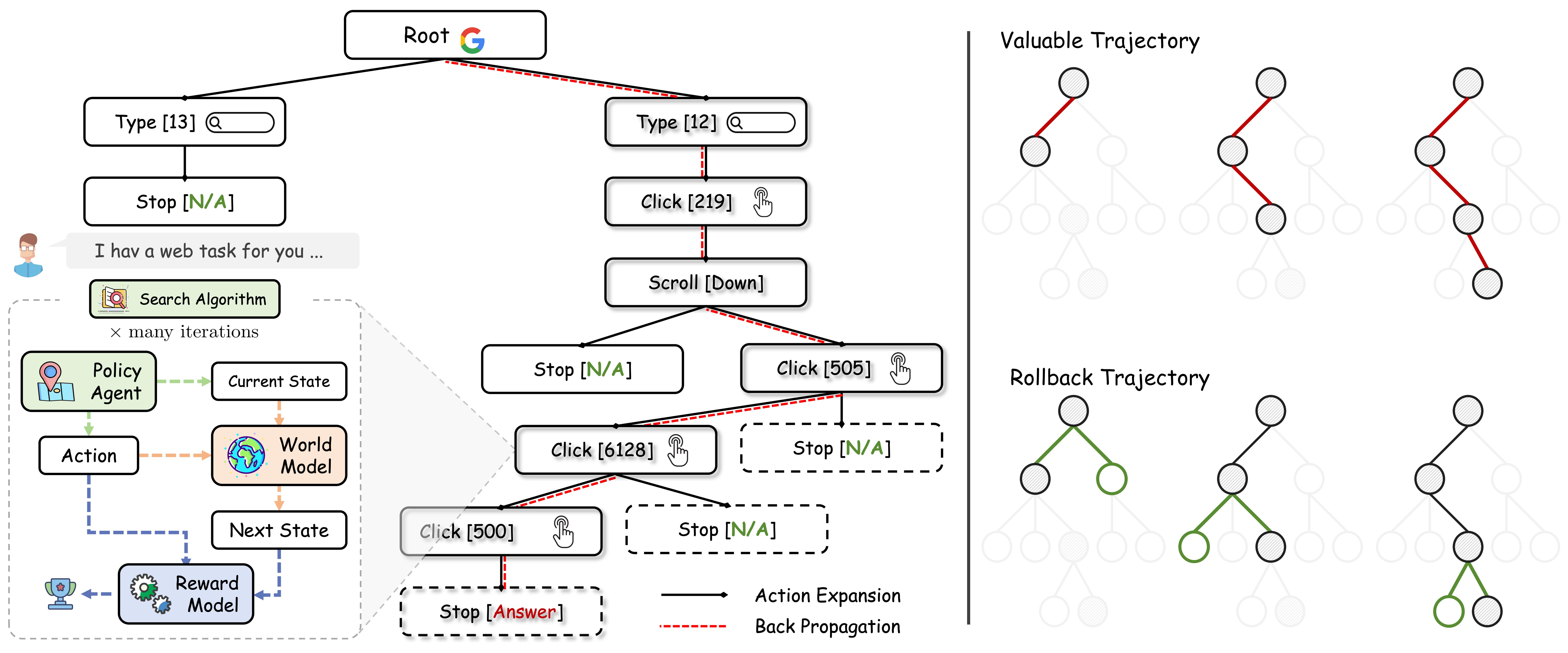}
    \caption{The pipeline of world model-guided Monte Carlo Tree Search and trajectory collection.}
    \label{fig:WebMCTS}
\end{figure*}

Consistent with other search-based approaches ~\cite{ReST-MCTS, Mulberry}, WebSynthesis requires a process reward model to guide the policy agent through step-by-step decision-making. However, web environments present unique challenges. If the reward model evaluates only the actions proposed by the policy agent without considering their broader context, the accuracy of action-level reward signals can be significantly decreased. Inspired by WMA \cite{WMA} and WebDreamer \cite{WebDreamer}, we leverage the world model to simulate the web environment and approximate the consequences of the agent's actions. This allows the process reward model to make more informed and accurate assessments. An overview of the framework is shown in Figure~\ref{fig:WebMCTS}. We now define the policy agent, the world model, and the process reward model as follows:
\begin{itemize}[leftmargin=1em]
    \item \textbf{Policy Agent}: At each time step $t$, a policy agent $\pi$ first generates an action $a_{t}$ according to the previous observation $o_{t-1}$ and user instruction $\text{q}$. Please refer to Table \ref{tab:plicy agent} for prompt details.
    \item \textbf{World Model}: A web world model $\omega$ then predicts the next observation $o_{t}$ based on $o_{t-1}$ and $a_{t}$. Please refer to Table \ref{tab:world model} for prompt details.
    \item \textbf{Process Reward Model}: A reward model $\gamma$ evaluates the actions issued by the policy agent based on the $o_{t}$ and the user's intent $\text{q}$. Please refer to Table \ref{tab:reward model} for prompt details.
\end{itemize}
Based on the above definition, the entire target of web navigation is formulated as follows: 

\begin{equation}\label{eq:1}
    \text{argmax}_{ \{a_{0},a_{1},\cdots a_{T} \} } \;\sum_{t=0}^{T} \gamma_{\theta}(o_{t}, a_{t})
\end{equation}

where $o_{t} \sim \omega_{\theta}(o_{t} \vert o_{t-1},a_{t})$, $a_{t} \sim \mathcal{\pi_{\theta}}(a_{t} \vert o_{t-1},\text{q})$ and $\text{argmax}_{ \{a_{0},a_{1},\cdots a_{T} \} }$ refers to the search algorithm. Within a fixed time $T$ or step budget, the objective of the search is to identify a trajectory that maximizes the reward.

According to Eq.~\ref{eq:1}, we employ the process reward model $\gamma$ to evaluate the quality of partial trajectories, enabling both selection and backpropagation at intermediate nodes. Building on this foundation, we formulate the world model-guided Monte Carlo Tree Search (\textbf{WebMCTS}) as an iterative algorithm comprising three main stages: node selection, action expansion, and backpropagation. In each iteration, the search tree is expanded based on candidate actions proposed by the policy agent $\pi$. Each node in the tree represents a specific action and stores the predicted next-state observation $o_{t+1}$ from the world model, the corresponding reward feedback $v_C$ from the reward model, and the number of visits $n_C$.

\textbf{Node Selection}: Following standard MCTS settings~\cite{ReST-MCTS, MCTS-survey}, node selection is guided by the Upper Confidence Bound (UCB) strategy, which balances exploration and exploitation during the search process. The UCB score for a child node $\text{C}$ is defined as: $$\text{U}_{C} = v_{C} + \epsilon\cdot \sqrt{\dfrac{\ln{n_{P}}}{n_{C}}}$$where $n_P$ is the number of visits to the parent node, and $\epsilon$ is an exploration constant. At each iteration, the algorithm selects the child node with the highest UCB score to continue the search.

\textbf{Action Expansion}: Given the node $\text{C}$ selected by the UCB strategy, we sample candidate actions based on the state associated with $\text{C}$. To ensure sufficient exploration and maintain the breadth of the search tree, WebMCTS samples at least three distinct potential actions at each expansion step. Additionally, since different branches of the search tree may reach the same web page (i.e., the same URL) at different points during the search, we adopt a caching mechanism to ensure consistency in the generated states from the world model. Specifically, we maintain a hash table that uses the URL as the key to store previously generated page states. When the same URL is encountered again, the cached version is reused, thereby avoiding duplicate state generation and improving both consistency and efficiency during tree expansion.

\textbf{Backpropagation}: Finally, we start value backpropagation from the selected node $\text{C}$ and update the values of its ancestor nodes using a weighted average strategy. Specifically, for each visited node $\text{C}$, we update its visit count and value estimate as follows: $$n_{C} \leftarrow n_{C}+1, \; \text{and} \;\, v_{C} \leftarrow \dfrac{\sum_{i} n_{C_{i}} \cdot v_{C_{i}} }{\sum_{i} n_{C_{i}}}$$After multiple iterations, this process produces a web action tree simulated by the world model, capturing a diverse set of possible interaction trajectories.

\begin{figure*}[!tp]
    \centering
    \includegraphics[width=1\linewidth]{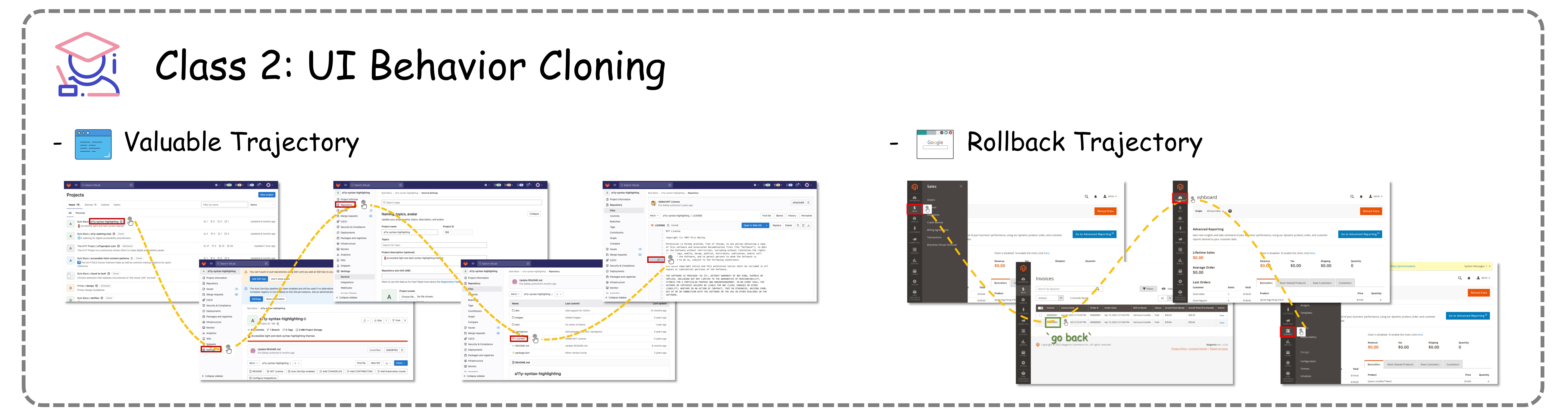}
    \caption{An overview of the UI Behavior Cloning dataset, which provides multi-step demonstrations for training policy agents. Valuable trajectories capture high-quality, target-directed browsing sessions, while rollback trajectories illustrate error recovery through explicit "go\_back" corrections. These datasets bridge the gap between understanding individual UI elements and performing end-to-end web navigation tasks.}
    \label{fig:class2}
\end{figure*}

\subsection{Stage 3: Trajectory Collection}\label{sec3.3}
Building on Stage 2, the resulting web action tree contains not only the trajectory $\tau = \{o_{t}, a_{t}\}_{t=0}^{N}$ that successfully accomplishes the task, but also a diverse set of meaningful yet unsuccessful attempts encountered during the search for $\tau$. In this subsection, we describe the process of extracting training trajectories from the constructed action tree. Our objective is to identify and categorize two types of trajectories, as illustrated in Figure~\ref{fig:class2}, (1) valuable trajectories, which positively contribute to policy learning, and (2) rollback trajectories, which represent failed exploration paths that can be used to improve the agent’s robustness and error-awareness.

\textbf{Valuable Trajectory}: Given that the initial policy agent may generate redundant or semantically meaningless actions when interacting with the web world model, we first prune the action tree before collecting training trajectories. We leverage GPT-4 to detect nodes containing repeated or overly similar actions. For example, in type-related actions, the agent may issue both "type [1201] [bus stop near CMU] [1]" and "type [1201] [bus stop near Carnegie Mellon University] [1]", which result in nearly identical state predictions from the web world model. In these cases, we remove the redundant nodes and merge their child nodes into their respective parent nodes. 

We employ the node value $v_{C}$ as a criterion for identifying valuable trajectories, where higher values signify a greater degree of task completion in the current state. Specifically, we perform a depth-first search (DFS) over the tree to locate target nodes whose values exceed a predefined threshold. For each identified target node, we trace the path backward to the root node, thereby extracting the corresponding trajectory for training.

\textbf{Rollback Trajectory}: The core idea of rollback is to enable the policy agent to dynamically assess the effectiveness of its actions and their outcomes at each step along the trajectory. When an action leads to a result that deviates from the user’s intent, the agent is expected to autonomously recognize the negative consequence of the incorrect decision, revert to a previously valid state $o_{t-1}$, and resume the reasoning process from that point. In WebMCTS, the resulted action tree maintains multiple parallel branches, which allows the agent to identify a diverse set of rollback trajectories from different high-value nodes within a single search episode. This contrasts with traditional linear or single-path rollback strategies, which typically generate only one trajectory per run and risk overlooking alternative valuable paths.

To extract rollback trajectories, we operate on each intermediate node $C$ within a valuable trajectory. Specifically, we begin by identifying the set of unsuccessful sibling nodes $S$ of $C$, along with their common ancestor node $P$. For each node in $S$, we leverage GPT-4 to synthesize a reflection that traces back to $P$, identifying the appropriate corrective action, denoted as "$\text{go\_back}$". We then revise the historical action sequence at $P$ to incorporate this correction, and construct a new trajectory that extends from $P$ through $S$. This process yields rollback trajectories in the form ${S \rightarrow P \rightarrow C}$, capturing alternative reasoning paths that arise from failed attempts and their corresponding corrective feedback.

\subsection{Stage 4: Policy Agent Training}\label{sec3.4}

As shown in Table \ref{tab: CL}, the training of the policy agent consists of two main components: UI fundamental capabilities and UI behavior cloning, which correspond to the training data collected in Stage 1 and Stage 3, respectively.

\begin{table*}[!t]
    \centering
    \caption{Curriculum overview for WebSynthesis training. Each class ('C') are designed for a specific capability, with approximate dataset scale and core learning objective.}
    \vspace{0.2cm}
    \renewcommand\arraystretch{1.3}{
        \rowcolors{3}{}{white}  
        \resizebox{1.0\linewidth}{!}{
            \begin{tabular}{c c c c}
                \toprule
                \textbf{Lessons} & \textbf{Category} & \textbf{Scale} & \textbf{Core Goal}\\
                \midrule
                C1\&stage 1 & Dense Captioning      & 2k & Grasp the overall semantics of TextUI expressions \\
                C1\&stage 2 & Element Functionality & 6k & Refine its fine-grained understanding of specific UI functions \\
                C1\&stage 3 & State Transition Perdition      & 7k & Strengthen its ability to predict page-state transitions \\
                \hline
                \rowcolor{gray!15} C2. Behavior Cloning & Valuable \& Rollback Trajectories  & 4k & Perform SFT on both valuable and rollback trajectories \\
                \bottomrule
            \end{tabular}
        }
    }
    \vspace{-0.2cm}
    \label{tab: CL} 
\end{table*}

\textbf{UI Fundamental Capabilities}: The first step is Supervised Finetuning (SFT) on the collected data in stage 1, which enhances the fundamental UI understanding capability. Since the Stage 1 dataset spans multiple dimensions, some designed to improve the model’s understanding of local and global UI information, and others intended to simulate simple user intentions, it is inappropriate to train on all samples at the same time. We thus adopt a curriculum learning (CL) strategy that mimics the human learning process, encouraging the model to first learn from simpler tasks and gradually progress to more complex ones. To enhance data quality, we follow InfiGUIAgent \cite{InfiGUIAgent} and implement instruction enhancement and response refinement process: (1) for data of different dimensions, we prompt GPT-4 to generate various instruction templates to enhance the logic between instructions and answers and; (2) for data with complex or inconsistent response formats, we leverage GPT-4 to reformulate responses while preserving their semantic content.

\textbf{UI Behavior Cloning}: Despite the above fundamental capabilities, the policy agent still lacks the ability to perform full web navigation due to the absence of complete trajectory training. To address this, we  adopt supervised fine-tuning using both valuable and rollback trajectories, following a procedure consistent with existing methods based on manually collected trajectory data. After the above training procedure, the policy agent acquires the essential capabilities required to complete web browsing tasks and can autonomously execute operations based on user instructions.


\section{Experiments}\label{sec4}

\subsection{Experimental Settings}\label{sec4.1}

\textbf{Model Settings}: We employ Qwen2.5-7B-Instruct as the base model for the policy agent and web world model training, and GPT-4 as the process reward model for tree search. We construct our dataset in online setting using the provided web environment for world model training. All training is performed as low rank adaption (LoRA) fine-tuning. Details are available in the Appendix.

\begin{table}[!t]
  \centering
  \small
  \setlength{\tabcolsep}{8pt}
  \renewcommand{\arraystretch}{1.03}
  \caption{Evaluation results (\%) on different WebArena subsets.  
  \textcolor{darkred}{\textbf{Red}} indicates the best score in each column; \underline{underlined} numbers are second best. We also report the task success rate within one and three attempts, where a task is considered successfully completed if at least one of the sampled trials passes the evaluation criteria.}
  \begin{tabular}{cc|cccccc}
        \toprule
        \textbf{Model} & \textbf{Param Size} & \textbf{Shopping} & \textbf{Admin} & \textbf{Reddit} & \textbf{Gitlab} & \textbf{Maps} & \textbf{Overall} \\
        \midrule
        Qwen2.5 & 7B & 2.17 & 0.00 & 0.00 & 6.25 & - & 2.24 \\
        LLaMA3.1 & 8B & 8.70 & 5.71 & 0.00 & 6.25 & - & 5.97 \\
        GPT-4 (CoT) & N/A & 13.04 & 17.14 & 9.52 & \textcolor{darkred}{\textbf{18.75}} & 7.14 & 13.58 \\
        \midrule
        \rowcolor{gray!20}
        \textbf{\textit{Pass@1}} &  &  &  &  &  &  &  \\
        AgentTrek \cite{AgentTrek} & 7B & 15.22 & 11.43 & 4.76 & 3.13 & - & 9.70 \\
        OS-Genesis \cite{OS-Genesis} & 7B & 10.87 & 14.29 & 0.00 & \underline{15.63} & - & 11.19 \\
        \textbf{WebSynthesis} & 7B & \underline{19.57} & \underline{17.14} & \underline{9.52} & 9.38 & - & \underline{14.93} \\
        \midrule
        \rowcolor{gray!20}
        \textbf{\textit{Pass@3}} &  &  &  &  &  &  &  \\
        AgentTrek \cite{AgentTrek} & 7B & 19.57 & 11.43 & 9.52 & 3.13 & - & 11.94 \\
        OS-Genesis \cite{OS-Genesis} & 7B & 19.57 & \textcolor{darkred}{\textbf{31.43}} & 0.00 & 15.63 & - & 18.66 \\
        \textbf{WebSynthesis} & 7B & \textcolor{darkred}{\textbf{28.26}} & 20.00 & \textcolor{darkred}{\textbf{14.29}} & 12.50 & - & \textcolor{darkred}{\textbf{20.15}} \\
        \bottomrule
\end{tabular}
  \label{tab:webarena_results}
\end{table}

\begin{table}[!t]
  \centering
  \small
  \setlength{\tabcolsep}{8pt}
  \renewcommand{\arraystretch}{1.05}
  \caption{Evaluation on WebArena with and w/o TextUI warm-up. The value inside the bracket denotes the absolute change relative to the same model \emph{without} TextUI: improvements are highlighted in \textbf{\textcolor{darkred}{red}} with a “\(+\)” sign, whereas degradations are highlighted in \textbf{\textcolor{darkgreen}{green}} with a “\(-\)” sign.}
  \begin{tabular}{ccccccc}
    \toprule
    \textbf{Model} & \textbf{Shopping} & \textbf{Admin} & \textbf{Reddit} & \textbf{Gitlab} & \textbf{Maps} & \textbf{Overall} \\ \midrule

    AgentTrek-7B    & 19.57\inc{4.35} & 14.29\inc{2.86} & 4.76 & 3.13 & -- & 11.94\inc{2.24} \\ 
    
    OS-Genesis-7B   & 17.39\inc{6.52} & 14.29 & 19.05\inc{19.05} & 6.25\dec{9.38} & -- & 14.93\inc{3.74} \\ 

    \rowcolor{lightrow}
    WebSynthesis-7B & 19.57\inc{6.53} & 17.14\inc{5.71} & 9.52\inc{9.52} & 9.38 & -- & 14.93\inc{5.23} \\ 
    
    \bottomrule
  \end{tabular}
  \label{tab:textui-warmup}
\end{table}

\textbf{Evaluation Benchmarks}: For the web navigation task, WebArena serves as a dynamic benchmark for evaluating autonomous agents. It covers domains such as maps, e-commerce (Shopping), content management (CMS), social forums (Reddit), and software development (Gitlab). Policy agents are required to make sequential decisions starting from an initial state, and task success is evaluated via string matching or url matching between the agent’s final output and the ground-truth answer. We follow the standard WebArena setup and adopt the following action space, including "click", "type", "hover", "scroll", "goto", "go\_back", and "stop". In the original WebArena environment, it comprises 812 challenging tasks derived from 241 task templates. Considering the cost of evaluation, we use 165 test cases from WebArena-Lite \cite{VisualAgentBench} for evaluation. Observations of the web environment are represented by the accessibility tree.

\textbf{Baseline Settings}: We compare our approach with the following baselines: (1) Qwen2.5-7B-Instruct, LLaMA3.1-8B-Instruct, and GPT-4, all evaluated under a setting with chain-of-thought prompting~\cite{WebArena}; (2) OS-Genesis~\cite{OS-Genesis}, which utilizes 7.4k real trajectories collected from WebArena; (3) AgentTrek \cite{AgentTrek}, which generates web agent trajectories by leveraging publicly available tutorials and contains 57k trajectory records. To ensure a fair comparison of headless browsing capabilities, we use only the text-based state representation (i.e., A11y) from the trajectories as input for all methods. For AgentTrek, we sample 20k trajectories from the original dataset to align with the action space used in WebArena. Both OS-Genesis and AgentTrek are fine-tuned on Qwen2.5-7B-Instruct. Detailed training settings are provided in the Appendix.

\subsection{Main Results}\label{sec4.2}

Table~\ref{tab:webarena_results} presents that Qwen2.5-7B, training with WebSynthesis data, leads to notable performance improvements. Compared with chain-of-thought (CoT) prompting-based methods, including open-source models such as Qwen2.5-7B-Instruct and LLaMA3.1-8B-Instruct, as well as closed-source models like GPT-4, WebSynthesis demonstrate superior performance across multiple WebArena subsets. Notably, WebSynthesis-SFT achieves a higher overall score (14.93\%) than GPT-4 (13.58\%), while significantly outperforming Qwen2.5-7B-Instruct (2.24\%).

In comparison to methods trained on real world trajectories, such as OS-Genesis-7B, which is trained on 7.4k real samples, WebSynthesis still yield better results. For example, WebSynthesis outperforms OS-Genesis by +3.74\% overall in Pass@1, despite being trained on a substantially smaller dataset (approximately 4k samples). These results suggest that high-quality synthetic trajectories guided by WebMCTS are comparably or more effective, even at a smaller scale. 

\subsection{Analysis}\label{sec4.3}

\subsubsection{How TextUI Works?}

A core step of WebSynthesis is a preliminary TextUI warmup phase conducted before formal trajectory-level training. This phase is designed to equip the policy agent with fundamental UI understanding capabilities. To verify the effectiveness of TextUI warm-up in enhancing trajectory learning, we compare and analyze the performance of three trajectory learning methods, including OS-Genesis, AgentTrek and WebSynthesis, with and without TextUI warmup. 

The experimental results are presented in Table \ref{tab:textui-warmup}. It is obvious that models with TextUI adaptation prior to trajectory-level fine-tuning consistently outperform their counterparts trained without such warmup. In particular, the improvement for OS-Genesis exceeds its original performance by 33.4\%. This showcases that fine-tuning on real-world collected trajectories is not a one-shot solution. Without sufficient familiarity with the structure and layout of text-based UI, trajectory-level training alone is unlikely to yield the desired performance.

\begin{table}[!t]
    \footnotesize
    \setlength{\tabcolsep}{6.45pt}
    \renewcommand{\arraystretch}{1.1}
    \caption{Ablation study on the contribution of \textit{UI Fundamental Capabilities} (FC.) and \textit{UI Behavior Cloning} (BC.) tasks. "Cap." stands for dense captioning, "Func." represents element functionality, and “Trans.” refers to state transition prediction. The trajectories $\tau_{\text{val}}$ and $\tau_{\text{roll}}$ denote the valuable and rollback trajectories respectively. A $\checkmark$ indicates that the task the corresponding task type was included during training. The \textcolor{gray!75}{shaded} row corresponds to our final method, \textbf{WebSynthesis}.}
    \begin{tabular}{ccc|cc|cccccc}
        \toprule
        \multicolumn{3}{c|}{\textbf{FC.}} & \multicolumn{2}{c|}{\textbf{BC.}} & \multirow{2}{*}{\textbf{Shopping}} & \multirow{2}{*}{\textbf{Admin}} & \multirow{2}{*}{\textbf{Reddit}} & \multirow{2}{*}{\textbf{Gitlab}} &  \multirow{2}{*}{\textbf{Maps}} & \multirow{2}{*}{\textbf{Overall}} \\
        \cmidrule(lr){1-3} \cmidrule(lr){4-5} 
        Cap. & Func. & Trans. & \textbf{$\tau_{roll}$} & \textbf{$\tau_{val}$} & & & & & & \\
        \midrule
        & & & $\checkmark$ & & 4.35 & 0.00 & 0.00 & 0.00 & - & 1.49 \\
        & & & & $\checkmark$ & 6.52 & 2.86 & 9.52 & 6.25 & - & 5.97 \\
        & & & $\checkmark$ & $\checkmark$ & 13.04 & 11.43 & 0.00 & 9.38 & - & 9.70 \\
        \midrule
        $\checkmark$ & & & $\checkmark$ & $\checkmark$ & 17.39 & 11.43 & 9.52 & 0.00 & - & 10.45 \\
        $\checkmark$ & $\checkmark$ & & $\checkmark$ & $\checkmark$ & \textbf{21.74} & 14.29 & 4.76 & 3.13 & - & 12.69 \\
        \midrule
        \rowcolor{gray!20}
        $\checkmark$ & $\checkmark$ & $\checkmark$ & $\checkmark$ & $\checkmark$ & 19.57 & \textbf{17.14} & \textbf{9.52} & \textbf{9.38} & - & \textbf{14.93} \\
        \bottomrule
    \end{tabular}
    \label{tab:Ablation}
\end{table}

\subsubsection{Ablation Studies of Fundamental Understanding}\label{sec4.3.1}
In the UI fundamental capabilities training phase, tasks from three dimensions are incorporated into the curriculum learning framework (see Section~\ref{sec3.1}). To further investigate which task type plays a critical role, we conduct sequential ablation experiments by incrementally adding tasks according to their difficulty levels. As shown in Table~\ref{tab:Ablation}, the three-stage curriculum learning gradually improved the web navigation ability of the policy agent. In particular, after the introduction of the state transition task in the third stage, the overall performance was improved by 5.23\% compared with the first two stages, indicating that the transition process of modeling the world state has a significant promoting effect on policy learning. This observation verifies the necessity of guiding the model to understand the dynamics of UI changes in the early stage of training, and provides a more accurate foundation for environmental modeling for subsequent multi-step reasoning and action execution.

In addition, from the perspective of task dimension, although the introduction of dense captioning or functionality description can bring some improvement, its gain is relatively limited. In contrast, the combination of tasks (caption + functionality + transition) has a more significant improvement in policy performance, showing the multi-dimensional of UI understanding ability. The gray area in the last row shows the complete WebSynthesis method, which achieved the highest performance in each task scenario, verifying the effectiveness of the curriculum learning strategy.

\subsubsection{Ablation Studies of Behaviors Cloning}\label{sec4.3.2}
Table~\ref{tab:Ablation} presents the ablation results comparing different trajectory supervision strategies. We examine four configurations: training on only valuable trajectories ($\tau_{\text{val}}$), only rollback trajectories ($\tau_{\text{roll}}$), their combination ($\tau_{\text{val}} \cup \tau_{\text{roll}}$), and full WebSynthesis training, which incorporates TextUI warm-up followed by combined trajectory training.

We observe that training solely on rollback trajectories ($\tau_{\text{roll}}$) leads to the weakest performance (1.49\% overall). This suggests that without exposure to successful trajectories, the agent becomes overly cautious and prone to issuing go\_back actions prematurely, ultimately losing the ability to explore and complete tasks. This result highlights the importance of learning from target-directed behavior to establish confident forward execution.

Training on valuable trajectories only ($\tau_{\text{val}}$) results in moderate performance (5.97\% overall), as it teaches the agent to reach task goals but fails to expose it to real-world error patterns or recovery strategies. When both $\tau_{\text{val}}$ and $\tau_{\text{roll}}$ are used together, the model achieves a significant improvement (9.70\% overall), indicating that exposure to both successful demonstrations and failure recovery paths enables the agent to reason more robustly in uncertain situations.

Importantly, WebSynthesis, which adds TextUI warm-up before trajectory-level fine-tuning, achieves the best overall performance (14.93\%). This highlights the importance of the UI fundamental capabilities in improving the model’s familiarity with the structure and layout of text-based UIs, thus facilitating more effective and sample-efficient policy learning in subsequent training stages. Without this foundation, the agent may struggle to interpret the UI context even with well-designed trajectories.


\subsubsection{How Data Scaling Helps Agent Abilities?}

\begin{wrapfigure}{r}{0.5\textwidth}  
    \centering
    \includegraphics[width=0.5\textwidth]{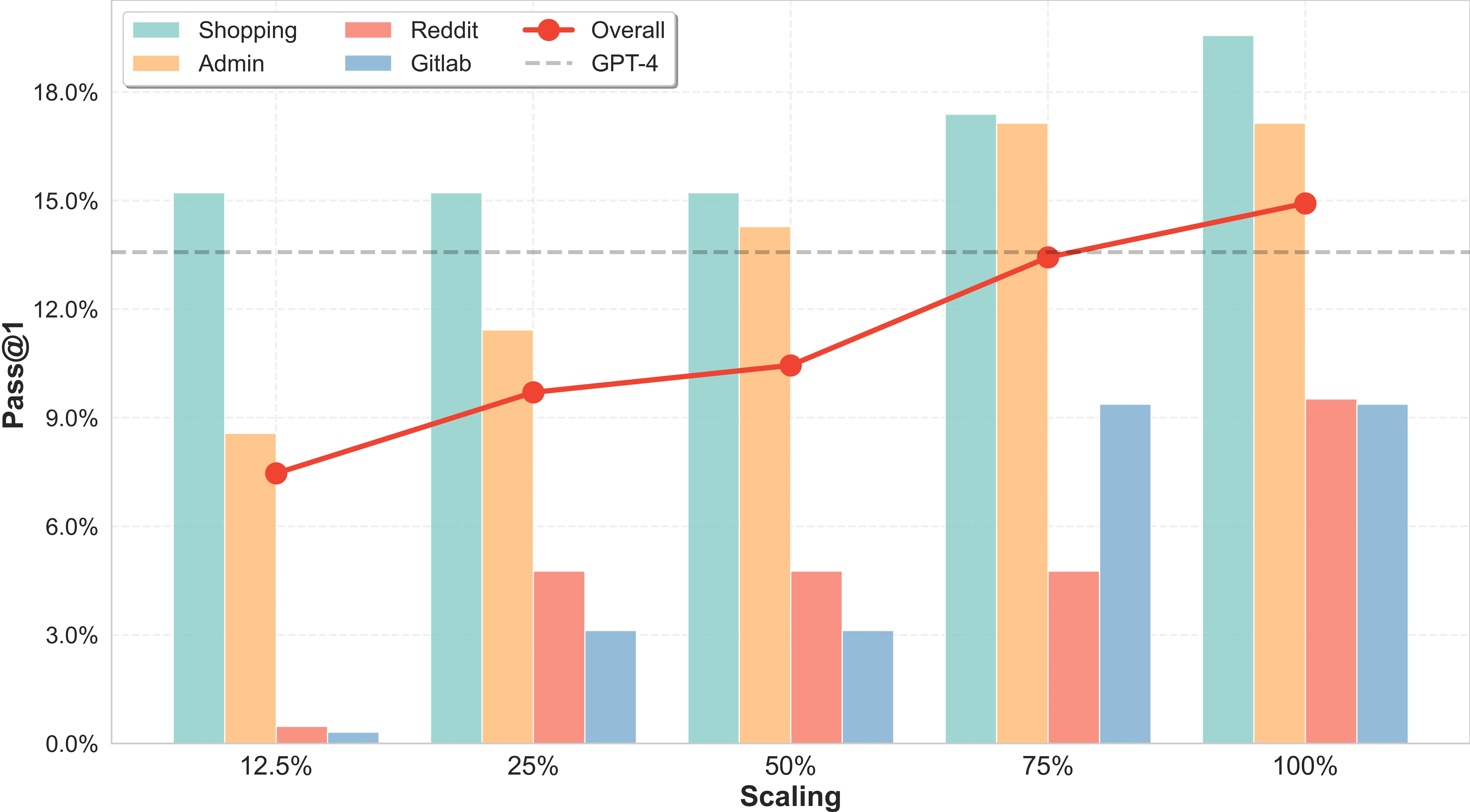}  
    \caption{Performance improvement with the synthetic trajectory data scaling up}  
\end{wrapfigure}

By controlling the proportion of the WebSynthesis dataset, we aimed to explore the impact of increasing the synthetic data scale on the performance of policy agents. As shown in the Figure, our evaluation on WebArena demonstrates a steady improvement in performance with the expansion of data scale. Notably, the performance gain is 7.47\% as the data scale increases from 12.5\% (approximately 500 samples) to 100\%. Remarkably, at 75\% of the data (approximately 4k samples), the performance already matches that of GPT-4. This suggests that the high-information-density data synthesized by WebMCTS can achieve significant sample efficiency improvements with a smaller dataset. This finding highlights that automated synthetic data generation is a viable strategy to bridge the performance gap with real-world collected data, underscoring the tremendous potential for future scalability.

\section{Conclusion}\label{sec5}
In this work, we propose WebSynthesis, a framework that integrates world model learning with Monte Carlo Tree Search (MCTS) to significantly reduce the online cost of synthesizing high-quality Web UI trajectories. Through a two-stage curriculum, comprising UI fundamental understanding and UI behavior cloning, the policy agent acquires web navigation capabilities. Notably, the agent trained with WebSynthesis on a small-scale synthetic dataset achieves performance comparable to, or even surpassing, that of models trained on large-scale real-world data.

While WebSynthesis demonstrates the potential of leveraging world models to replace real web environments for offline data collection, a more ambitious and forward-looking direction is to integrate world models into online reinforcement learning, following paradigms explored in real-world RL research~\cite{Muzero, ZeroSearch, Embodied-R, Autonomous-Driving-Survey}. In such a setting, the policy agent can continuously improve its planning and decision-making capabilities by directly interacting with a simulated environment constructed by the world model.

However, several key challenges remain before this vision can be fully realized in web environments. First, unlike board games or robotic control tasks, web interfaces are highly diverse, partially observable, and non-stationary, making it difficult to learn a stable and generalizable world model. Second, compounding errors in multi-step rollouts can significantly degrade performance, particularly when the agent relies heavily on imagined trajectories for exploration. Third, integrating model-based environments into online, closed-loop settings remains an open research problem. It is crucial not only to prevent model collapse over time, but also to ensure that the agent's behavior aligns with human-aligned criteria. Achieving this requires advances in both algorithmic frameworks and the fidelity of learned world models. Nevertheless, world models still hold strong potential for replacing real web environments in rare, costly, or safety-critical scenarios, where real-time interaction is limited or impractical.

\bibliographystyle{plain} 
\bibliography{neurips_2025}
\newpage







\appendix

\newpage
\section{UI Fundamental Understanding Instruction Template}
\begin{table*}[!ht]
\small
    \centering
    \begin{tcolorbox}[colframe=black, colback=gray!10!white, coltitle=black, boxrule=0.5mm]
    You're a helpful web GUI assistant.\\
    
    \textbf{OBSERVATION}: $\mathrm{\mathbf{\{Web\ State\}}}$ \\
    
    \textbf{Instruction}: Detail the main sections and functionalities available in this interface. \\
    
    \textbf{Output}: Let's think step-by-step. The web page displays a user interface from an e-commerce platform, specifically the "Pages" section. The layout is organized with a sidebar on the left featuring navigation icons for various functions such as Dashboard, Sales, Catalog, and more. The main area shows a search bar labeled Privacy Policy, indicating that users can search for specific pages. \\
    Below the search bar, there are options to manage actions and view filters. A table lists three pages with columns for ID, Title, URL Key, Layout, Store View Status, Created Date/Time Modified Date/Time Action. Each row provides detailed information about individual pages including their titles and statuses. \\
    This interface allows users to efficiently manage website content by enabling them to add new pages or modify existing ones through drop-down menus in the action column. Overall, it reflects a structured approach to page management within an e-commerce environment.
    \end{tcolorbox}
    \vspace{-0.2cm}
    \caption{Instruction template for $\mathrm{\mathbf{\{Dense\ Caption\}}}$.}
    \vspace{-0.2cm}
    \label{tab:textui_caption}
\end{table*}

\begin{table*}[!ht]
\small
    \centering
    \begin{tcolorbox}[colframe=black, colback=gray!10!white, coltitle=black, boxrule=0.5mm]
    You're a helpful web GUI assistant.\\
    
    \textbf{OBSERVATION}: $\mathrm{\mathbf{\{Web\ State\}}}$ \\
    
    \textbf{Instruction}: What job does the element [3230] accomplish? \\
    
    \textbf{Output}: Let's think step-by-step. The functionality of this element is to filter projects based on specific topics or themes. By selecting 'Topics', users can narrow down their search results to projects that match the chosen category, enhancing the relevance and discover ability of the projects displayed in the project list.
    \end{tcolorbox}
    \vspace{-0.2cm}
    \caption{Instruction template for $\mathrm{\mathbf{\{Functional\ Description\}}}$.}
    \vspace{-0.2cm}
    \label{tab:textui_function}
\end{table*}

\begin{table*}[!ht]
\small
    \centering
    \begin{tcolorbox}[colframe=black, colback=gray!10!white, coltitle=black, boxrule=0.5mm]
    You're a helpful web GUI assistant.\\
    
    \textbf{OBSERVATION}: $\mathrm{\mathbf{\{Web\ State\}}}$ \\
    
    \textbf{Instruction}: Predict how the UI evolves when this user interaction occurs: click [7716], where [7716] is 'type:link, text:Submit' \\
    
    \textbf{Output}: Let's think step-by-step. The user clicked on the "Submit" link, transitioning from a search results page to a submission creation page. This is evident from the removal of content related to search results, such as headings, articles, and links to individual posts, and the addition of elements specifically designed for creating submissions. The new page includes input fields for title and body, radio buttons for submission type (URL or image), a drop down menu for selecting a forum, and a submission button. The navigation bar remains consistent, but now supports submission-related functionalities.\\
    
    In summary, the webpage now displays a submission creation interface on Postmill. The top navigation bar includes links to "Postmill," "Forums," and "Wiki," along with a search bar and a "Submit" button. The main content area is structured for creating a submission, featuring input fields for Title (required), Body, and submission type options (URL or Image). Users can select a forum via a drop down menu labeled "Forum," which allows categorization of the submission. A button labeled "Create submission" is provided to finalize the process, and Markdown formatting is supported with help links available. This layout facilitates users in crafting and submitting content to the platform's forums."

    \end{tcolorbox}
    \vspace{-0.2cm}
    \caption{Instruction template for $\mathrm{\mathbf{\{State\ Transmission\}}}$.}
    \vspace{-0.2cm}
    \label{tab:textui_transmission}
\end{table*}

\section{Prompt Template for Policy Agent, World Model and Reward Model}
\begin{table*}[!ht]
\small
    \centering
    \begin{tcolorbox}[colframe=black, colback=gray!10!white, coltitle=black, boxrule=0.5mm]

    You are an autonomous intelligent agent tasked with navigating a web browser. You will be given web-based tasks. These tasks will be accomplished through the use of specific actions you can issue. \\
    Here's the information you'll have:
    \begin{itemize}[leftmargin=1em]
        \item The user's \textbf{objective}: This is the task you're trying to complete. 
        \item The \textbf{current observation} of web page. This is a simplified representation of the webpage, refer to as accessibility tree (a11y), providing key information.
        \item The previous \textbf{trajectory}: This is the 'observations', 'thoughts' and 'actions' you have just performed. It may be helpful to track your progress. Each step is splited by <step></step> tag. 
    \end{itemize}
    
    \#\# \textbf{Action Space}\\
    The actions you can perform fall into several categories:
    
    \#\#\# \textbf{Page Operation Actions}:\\
    \text{click} [id]: This action clicks on an element with a specific id on the webpage.\\
    \text{type} [id] [content] [press\_enter\_after=0|1]: Use this to type the content into the field with id. By default, the "Enter" key is pressed after typing unless press\_enter\_after is set to 0.\\
    \text{hover} [id]: Hover over an element with id, this action may display hidden information about the element.\\
    \text{scroll} [direction=down|up]: Scroll the page up or down. This action will provide new or previously appeared page information.\\
    \#\#\#  \textbf{URL Navigation Actions}:\\
    \text{goto} [url]: Navigate to a specific url.\\
    \text{go\_back}: Navigate to the previously viewed page. \\
    \#\#\#  \textbf{Completion Action}:\\
    \text{stop} [answer]: Issue this action when you believe the task is complete. If the objective is to find a text-based answer, provide the answer in the bracket. If you believe the task is impossible to complete, provide the answer as "N/A" in the bracket.\\
    
    \#\#  \textbf{Tips}:
    \begin{itemize}[leftmargin=1em]
        \item If the page has element information that is useful for completing the task, you can perform page actions to explore, but please pay attention to the actual function of each element. It is forbidden to perform function B on an element that only has function A, such as retrieving other information besides geographic information in the search box of OpenStreetMap.
        \item Fuzzy search is prohibited. Your search must be based on a clear goal. For example, you can search for a pair of Nike shoes, but you are not allowed to search for a pair of shoes that cost around \$60.
        \item If the page doesn’t have information that helps you complete the task, you can perform url navigation actions, including the need to jump to a specific page (for example, jump to Reddit), or compare the information on the previous and next pages to help complete the task.
        \item If you think you have completed this task, please check your trajectory carefully and make a completion action carefully.
        \item If there is no information on the current page that can help complete the task, please also make a completion action carefully.
    \end{itemize}

    \#\# \textbf{Action Rules}:\\
    To be successful, it is very important to follow the following rules:
    \begin{enumerate}[leftmargin=1em]
        \item You should think step by step and then issue the next action. Start with a "Let's think step-by-step." phrase.
        \item You should only issue an action that is valid given the current web page.
        \item You should only issue one action at a time.
        \item Generate the action in the correct format. Start with a "In summary, the next action I will perform is" phrase, followed by action inside ``````. For example, "In summary, the next action I will perform is ```click [1234]```".
        \item Issue stop action when you think you have achieved the objective. Don't generate anything after stop.
    \end{enumerate}
    
    \end{tcolorbox}
    \vspace{-0.2cm}
    \caption{Policy Agent Prompt Template.}
    \vspace{-0.2cm}
    \label{tab:plicy agent}
\end{table*}

\begin{table*}[!ht]
\small
    \centering
    \begin{tcolorbox}[colframe=black, colback=gray!10!white, coltitle=black, boxrule=0.5mm]

    You are an autonomous intelligent agent tasked with navigating a web browser. You will be given a web GUI-based task. Specifically, you need to predict the next web page observation based on the current observation of the web browser and the given action.\\
    Here's the information you'll have:
    \begin{itemize}[leftmargin=1em]
        \item The current web \textbf{page observation}, which lists the IDs of all interactive elements on the current web page with their text content if any, in the format [id] [tagType] [text content]. tagType is the type of the element, such as button, link, or textbox. text content is the text content of the element. For example, [1234] [button] [’Add to Cart’] means that there is a button with id 1234 and text content ’Add to Cart’ on the current web page. [] [StaticText] [text] means that the element is of some text that is not interactive.
        \item The given \textbf{action}: This is the action you have already performed. The actions you have performed fall into several categories:
    \end{itemize}
    
    \#\# \textbf{Action Space}\\
    The actions you can perform fall into several categories:
    
    \#\#\# \textbf{Page Operation Actions}:\\
    \text{click} [id]: This action clicks on an element with a specific id on the webpage.\\
    \text{type} [id] [content] [press\_enter\_after=0|1]: Use this to type the content into the field with id. By default, the "Enter" key is pressed after typing unless press\_enter\_after is set to 0.\\
    \text{hover} [id]: Hover over an element with id, this action may display hidden information about the element.\\
    \text{scroll} [direction=down|up]: Scroll the page up or down. This action will provide new or previously appeared page information.\\
    \#\#\#  \textbf{URL Navigation Actions}:\\
    \text{goto} [url]: Navigate to a specific url.\\
    \text{go\_back}: Navigate to the previously viewed page. \\
    \#\#\#  \textbf{Completion Action}:\\
    \text{stop} [answer]: Issue this action when you believe the task is complete. If the objective is to find a text-based answer, provide the answer in the bracket. If you believe the task is impossible to complete, provide the answer as "N/A" in the bracket.\\
    
    *IMPORTANT*\\
    To be successful, it is very important to follow the following rules:
    \begin{enumerate}[leftmargin=1em]
        \item Please think step by step based on the current page observation and the actions taken, and give the maximum possible next page observation.
        \item You should ensure the richness of the observations of the web page to be predicted and support the continuous operation process.
        \item Please generate the content of the next page in the correct format. Start with the phrase "In summary, the next web page observation is" and then add supplements within ```<your generated contents>```. For example, "In summary, the next web page observation is ```Tab 0 (current): Projects ````".
        \item When you think you have achieved the full content prediction of the next page, issue a stop operation with [END]. Do not generate any content after stopping.""", 
    \end{enumerate}
    
    \end{tcolorbox}
    \vspace{-0.2cm}
    \caption{World Model Prompt Template.}
    \label{tab:world model}
\end{table*}

\begin{table*}[!ht]
\small
    \centering
    \begin{tcolorbox}[colframe=black, colback=gray!10!white, coltitle=black, boxrule=0.5mm]

    You are an expert in evaluating GUI agent task trajectories. Your task is to assess the quality and effectiveness of task trajectories for GUI manipulation tasks.\\
    A trajectory consists of the following components:
    \begin{enumerate}[leftmargin=1em]
        \item \textbf{User Instruction}: Describes the user’s intended task.
        \item \textbf{Action History}: Includes two key parts:
        \begin{itemize}[leftmargin=1em]
            \item[-] Reasoning and Action for Each Step: A sequence of actions performed by the agent, including the reasoning thought and final executed action.
            \item[-] The accessibility tree of the current web page: This is a simplified representation of the webpage, providing key information.
        \end{itemize}
    \end{enumerate}

    When evaluating a trajectory, consider these key aspects:\\    
    \end{tcolorbox}
\end{table*}

\begin{table*}[!t]
\small
    \centering
    \begin{tcolorbox}[colframe=black, colback=gray!10!white, coltitle=black, boxrule=0.5mm]

    \textbf{Evaluation Criteria}:\\
    1. Trajectory Coherence:
    \begin{itemize}[leftmargin=1em]
        \item[-] Do the steps and corresponding actions follow a logical sequence toward the goal?
        \item[-] Are the actions clearly described and specific?
        \item[-] Are there redundant or unnecessary actions?
    \end{itemize}
    2. Task Completion:
    \begin{itemize}[leftmargin=1em]
        \item[-] Does the trajectory successfully achieve the instructed task?
        \item[-] Are all necessary interactions completed?
        \item[-] Are error cases handled appropriately?
    \end{itemize}
    
    \textbf{Scoring Guidelines}:\\
    Rate the trajectory on a scale of 1 to 5 based on the evaluation criteria:
    \begin{itemize}[leftmargin=1em]
        \item[-] 5: The task is perfectly completed, successfully executing multiple actions to achieve the goal or return the correct answers. The sequence is logically clear with no noticeable redundancies.
        \item[-] 4: The task is mostly completed, successfully executing multiple actions. However, due to challenges or ambiguities in the instructions, the completion is not perfect, or there are inefficiencies in the process.
        \item[-] 3: The task is partially completed, with some successful actions executed. However, due to task or environmental constraints, the goal is not fully achieved, or the sequence ends in a loop or error.
        \item[-] 2: Only a few actions are executed. Although there is an attempt to complete the task, the trajectory deviates from the goal early on or demonstrates significant inefficiencies in execution and logic, e.g., repeat the same action.
        \item[-] 1: The task fails completely, with no meaningful actions executed at the start. The sequence either falls into an immediate deadlock, a repetitive loop, or demonstrates no value in completing the task.
    \end{itemize}

    Or the tasks are completely inaccessible.\\
    \textbf{Note}: If the task is relatively complex, but the trajectory demonstrates valuable attempts, even if the task is not fully completed, consider adjusting the score upward. However, if the task is complex but the trajectory fails to perform actions that contribute meaningfully to task completion, no extra points should be awarded. \\
    You need to judge the score based on the user instruction, agent’s actions and the current state of the webpage combined.\\
    \textbf{Response Format}:\\
    Format your response into two lines as shown below:\\
    Reason: <your thoughts and reasoning process for the score>\\
    Score: <your score from 1-5>
    \end{tcolorbox}
    \vspace{-0.2cm}
    \caption{Reward Model Prompt Template.}
    \vspace{-0.2cm}
    \label{tab:reward model}
\end{table*}

\begin{table*}[!ht]
\small
    \centering
    \begin{tcolorbox}[colframe=black, colback=gray!10!white, coltitle=black, boxrule=0.5mm]

    \textbf{OBJECTIVE}: \\
    <insert user objective here>\\

    \textbf{TRAJECTORY}: \\
    <step-$i$>
    \begin{itemize}[leftmargin=1em]
        \item[-] OBSERVATION: <insert historical observation here>
        \item[-] REASON FOR ACTION: <insert historical thought here>
        \item[-] ACTION: <insert historical action here>
    \end{itemize}
    </step-$i$>\\
    $\cdots$ \\
    
    \textbf{OBSERVATION}: \\
    <insert current observation here>\\
    
    What's the next action?
    
    \end{tcolorbox}
    \caption{Policy Agent Instruction Template.}
    \label{tab:The Trajectory Template.}
\end{table*}


\end{document}